\documentclass{article}

\PassOptionsToPackage{numbers, compress}{natbib}

\usepackage[preprint]{neurips_2023}

\usepackage[utf8]{inputenc} 
\usepackage[T1]{fontenc}    
\usepackage{hyperref}       
\usepackage{url}            
\usepackage{booktabs}       
\usepackage{amsfonts}       
\usepackage{nicefrac}       
\usepackage{microtype}      
\usepackage{xcolor}         
\usepackage[pdftex]{graphicx}
\usepackage{amsmath}
\usepackage{multirow}
\usepackage{tabularx}
\usepackage{todonotes}
\usepackage{caption} 
\usepackage{lineno}
\usepackage[frozencache,cachedir=.]{minted}
\usepackage{threeparttable}
\usepackage{footmisc}
\usepackage[title]{appendix}

\title{Driving with LLMs: Fusing Object-Level Vector Modality for Explainable Autonomous Driving}

\author{%
  Long Chen$^*$ \quad Oleg Sinavski$^*$ \quad Jan Hünermann \quad Alice Karnsund \\ \textbf{Andrew James Willmott} \quad \textbf{Danny Birch} \quad \textbf{Daniel Maund} \quad \textbf{Jamie Shotton} \\
  \\
  \textbf{Wayve} \\
  \texttt{research@wayve.ai}\\
  \texttt{$^*$equal contributions}\\
  }

\begin{document}

\maketitle
\thispagestyle{empty}
\pagestyle{empty}


\begin{abstract}
Large Language Models (LLMs) have shown promise in the autonomous driving sector, particularly in generalization and interpretability. We introduce a unique object-level multimodal LLM architecture that merges vectorized numeric modalities with a pre-trained LLM to improve context understanding in driving situations. We also present a new dataset of 160k QA pairs derived from 10k driving scenarios, paired with high quality control commands collected with RL agent and question answer pairs generated by teacher LLM (GPT-3.5). A distinct pretraining strategy is devised to align numeric vector modalities with static LLM representations using vector captioning language data. We also introduce an evaluation metric for Driving QA and demonstrate our LLM-driver's proficiency in interpreting driving scenarios, answering questions, and decision-making. Our findings highlight the potential of LLM-based driving action generation in comparison to traditional behavioral cloning. We make our benchmark, datasets, and model available \footnote{https://github.com/wayveai/Driving-with-LLMs\label{code}} for further exploration.
\end{abstract}

\section{Introduction}
\begin{figure}[hbt!]
    \centering
    \includegraphics[width=0.70\linewidth]{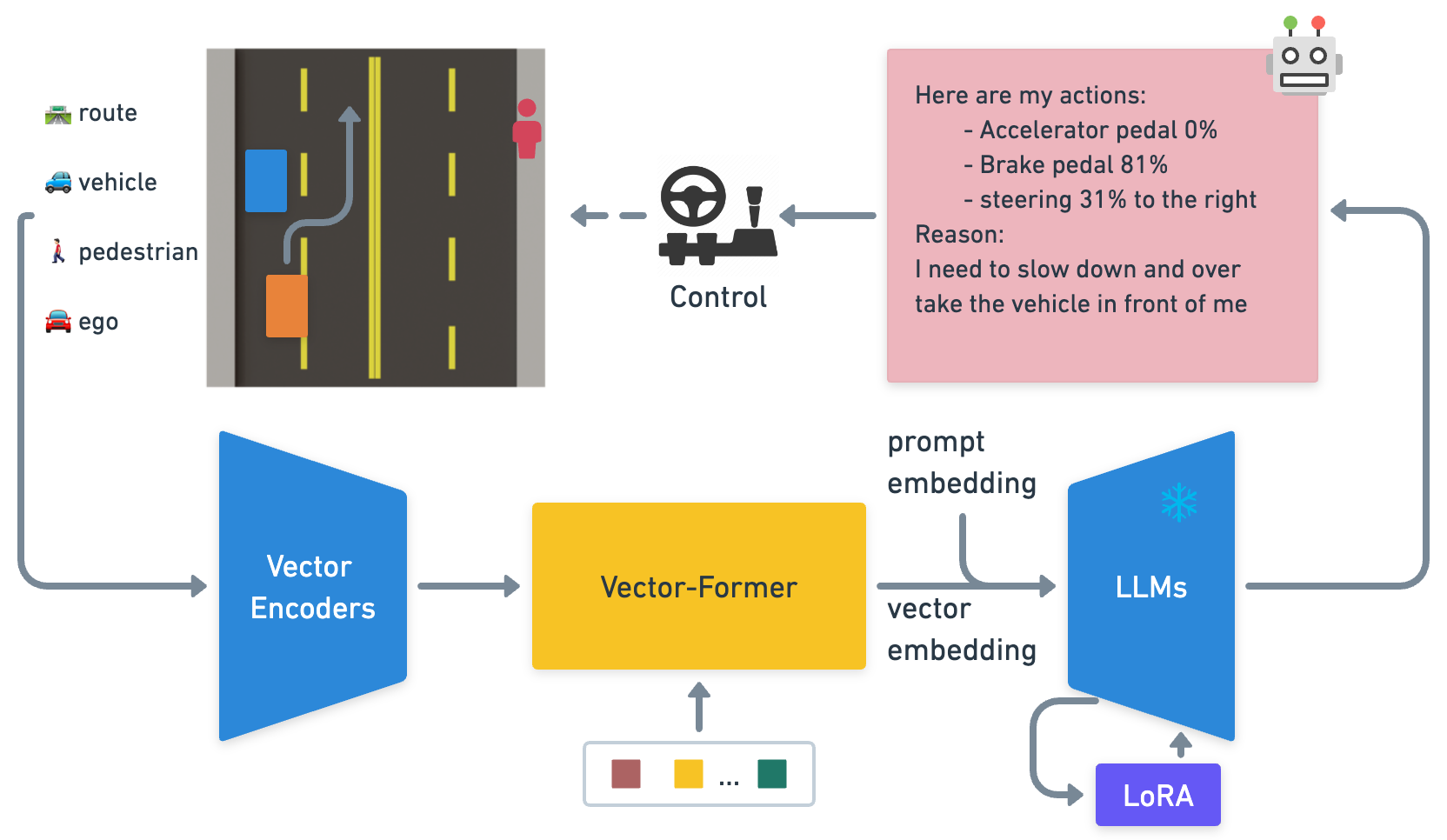}
    \caption{An overview of the architecture for Driving with LLMs, demonstrating how object-level vector input from our driving simulator is employed to predict actions via LLMs} 
    \label{fig:intro} 
\end{figure}
Remarkable abilities of Large Language Models (LLMs) demonstrate early signs of artificial general intelligence (AGI) \cite{bubeck2023sparks}, exhibiting capabilities such as out-of-distribution (OOD) reasoning, common sense understanding, knowledge retrieval, and the ability to naturally communicate these aspects with humans. These capabilities align well with the focus areas of autonomous driving and robotics \cite{wells2021explainable} \cite{lu2018robot}.

Modern scalable autonomous driving systems, whether they adopt an end-to-end approach using a single network \cite{hawke2021reimagining}, or a component-based configuration that combines learnable perception and motion planning modules \cite{chen2021data} \cite{https://doi.org/10.1002/rob.21918}, face common challenges. These systems often behave as 'black-boxes' in the decision making process, making it especially difficult to endow them with OOD reasoning and interpretability capabilities. Such issues persist even though there have been some strides towards addressing them \cite{Omeiza_2022}.

Textual or symbolic modality, with its inherent suitability for logical reasoning, knowledge retrieval, and human communication, serves as an excellent medium for harnessing the capabilities of LLMs \cite{rajani2019explain}. However, its linear sequential nature limits nuanced spatial understanding \cite{bubeck2023sparks}, a crucial aspect of autonomous navigation. Pioneering work in Visual Language Models (VLMs) has begun to bridge this gap by merging visual and text modalities 
\cite{wang2023largescale}, enabling spatial reasoning with the power of pre-trained LLMs. However, effectively incorporating the new modality into the language representation space requires extensive pretraining with a significant volume of labeled image data.

We propose a novel methodology for integrating the numeric vector modality, a type of data that is frequently used in robotics for representing speed, actuator positions and distance measurements, into pre-trained LLMs. Such modality is considerably more compact than vision alleviating some of the VLM scaling challenges. Specifically, we fuse vectorized object-level 2D scene representation, commonly used in autonomous driving, into a pre-trained LLM with adapters \cite{hu2021lora}. This fusion enables the model to directly interpret and reason about comprehensive driving situations. As a result, the LLMs are empowered to serve as the ``brain`` of the autonomous driving system, interacting directly with the simulator to facilitate reasoning and action prediction.

To obtain training data in a scalable way, we first use a custom 2D simulator and train a reinforcement learning (RL) agent to solve the driving scenarios, serving as a substitute for a human driving expert. To ground the object-level vector into LLMs, we introduce a language generator that translates this numerical data into textual descriptions for representation pretraining. We further leverage a teacher LLM (GPT) to generate a question-answering dataset conditioned on the language descriptions of 10k different driving scenarios. Our model first undergoes a pretraining phase that enhances the alignment between the numeric vector modality and the latent language representations. Next, we train our novel architecture to establish a robust baseline model, LLM-driver, for the driving action prediction and driving question answering tasks. We provide our datasets, evaluation benchmarks and a pre-trained model\footref{code} for reproducibility and hope to inspire and facilitate further advancements in the field. The subsequent sections of this paper detail the theoretical background, our proposed architecture and experimental setup, preliminary results, potential directions for future research, and implications of our work for the broader field of autonomous driving.

In this paper, we have made the following contributions:

\begin{enumerate}
    \item \textbf{Novel object-level multimodal LLM architecture:} We propose a novel architecture that fuses an object-level vectorized numeric modality into any LLMs with a two-stage pretraining and finetuning method.
    
    \item \textbf{Driving scenario QA task and a dataset:} We provide a 160k question-answer pairs dataset on 10k driving situations with control commands, collected with RL expert driving agents and an expert LLM-based question answer generator. Additionally, we also outline the methodology for further data collection.

    
    \item \textbf{Novel Driving QA (DQA) evaluation and a pretrained baseline:} We present a novel way to evaluate Driving QA performance using the same expert LLM grader. We provide initial evaluation results and a baseline using our end-to-end multimodal architecture. 
\end{enumerate}

Our work provides the first-of-its-kind baseline approach for integrating LLMs into driving task in simulation. This includes a comprehensive framework encompassing the simulator, automatic data collection, integration of a new object-level vector modality into LLMs, and the GPT-based evaluations approaches.  

\section{Related Works}
\subsection{End-to-End Autonomous Driving Systems}
There was a significant progress in end-to-end deep learning methods for autonomous systems in recent years  \cite{bansal2018chauffeurnet}, \cite{gao2020vectornet}, \cite{hawke2021reimagining}, with some of the earliest efforts dating back to ALVINN \cite{pomerleau1988alvinn} and more recent works such as \cite{NEURIPS2022_827cb489}. However, a fundamental challenge that remains with modern autonomous driving systems is the lack of interpretability in the decision making process \cite{BARREDOARRIETA202082}. Understanding why a decision is made is crucial for identifying areas of uncertainty, building trust, enabling effective human-AI collaboration and ensuring safety \cite{HumanInteractions}. We continue this line of research by adding vector/textual modality and pretrained LLMs to the end-to-end autonomous driving.

\subsection{Interpretability of Autonomous Driving Systems}
A variety of explainability methods have been introduced \cite{Holzinger2022} to understand the underlying decision process of deep neural networks. For example \cite{ribeiro2016should}, \cite{lundberg2017unified} and \cite{shrikumar2017learning} are well-established model-agnostic interpretability methods that generate explanations for individual predictions. Other methods such as gradient based \cite{selvaraju2017grad}, saliency maps \cite{simonyan2013deep} and attention maps \cite{xu2015show} target the inner operations of models to explain the decision making process. In the field of autonomous vehicles, visual attention maps, which highlight causally influential regions in driving images were proposed in \cite{kim2017interpretable}. In \cite{kim2018textual} the authors combined attention based methods with natural language to create an attention-based vehicle controller that provides natural language action descriptions and explanations based on a series of image frames. This work was further extended in \cite{kuhn2023textual}, where the authors improved the architecture by integrating part of speech prediction and special token penalties. Others argue that attention is not enough \cite{jain2019attention}, leading to multiple efforts to combine this approach with other explanatory methods. For example \cite{qiang2022attcat} proposes to explain transformers by leveraging attentive class activation tokens, encoded features, their gradients, and their attention weights simultaneously. Building on this research, we are proposing to use text modality for explainability in autonomous driving.



\subsection{Multi-modal LLMs in Driving Tasks}
Recently, there has been a notable trend towards integrating multiple modalities into unified large-scale models. Notable examples include VLMs such as \cite{radford2021learning}, \cite{alayrac2022flamingo}, \cite{li2023blip2}, and \cite{openai2023gpt4}, which effectively combine language and images to accomplish tasks like image captioning, visual question answering, and image-text similarity. Another noteworthy advancement \cite{girdhar2023imagebind} involves the fusion of information from six distinct modalities: text, image/video, audio, depth, thermal, and inertial measurements. This exciting development not only expands the possibilities for generating content using diverse data input and output types but also enables broader multi-modal search capabilities.

With camera sensors being one of the most common sensors used in autonomous driving \cite{s23063335}, a natural step to incorporate language has been through VLMs. For example \cite{roh2020conditional} uses images and language directions to train a driving policy. \cite{Kim_2020_CVPR} proposes a method for learning vehicle control with human assistance. The system learns to summarize its visual observations in natural language, predict an appropriate action response (e.g. “I see a pedestrian crossing, so I stop”), and predict the controls, accordingly. Using language to explain the inner workings of the model has also been explored in \cite{jin2023adapt}, where user-friendly natural language narrations and reasoning are provided for each decision making step of autonomous vehicular control and action.

In robotics, we have seen efforts fusing language with other modalities. Albeit outside of autonomous driving field, the closest work to ours \cite{roh2022languagerefer} utilizes point clouds with 3D bounding boxes of potential object candidates. It also uses a language utterance referring to a target object in the scene to train a model capable of identifying a target object from a set of potential candidates. Recently, the RT-2 paper \cite{rt22023arxiv} demonstrated a similar approach by utilizing LLMs for low-level robotics control tasks, including the joint training of VQA and control tasks. However, their framework is confined to the vision modality, whereas we introduce a novel methodology for grounding vector-based object level modalities into LLMs, facilitating interpretable control and driving QA tasks. In contrast to these existing efforts, the work presented in this paper is, to the best of our knowledge, the first to fuse numeric vector modality with language specifically in the domain of autonomous vehicles.

\section{Method}



\begin{figure*}[hbt]
    \centering
    \includegraphics[width=1.0\linewidth]{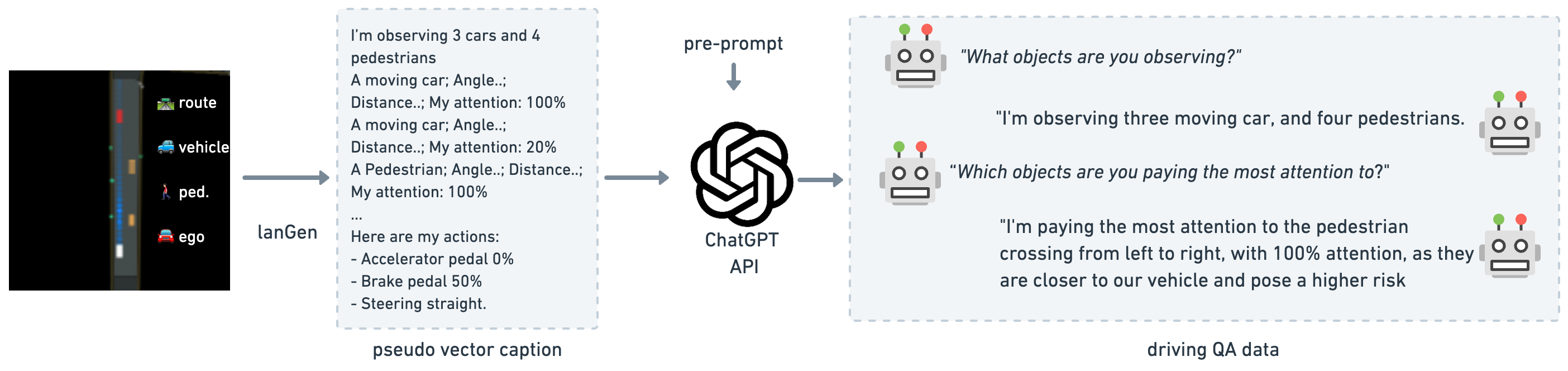}
    \caption{The illustration of our Driving QA Dataset automatic labelling process} 
    \label{fig:arc} 
\end{figure*}

\subsection{Data Collection using RL experts}
To generate language-based grounded driving datasets, we use a custom-built realistic 2D simulator with procedural generation of driving scenarios. We use an RL agent that solves the simulated scenarios using an object-level ground-truth representation of the driving scene. In our approach, we map a vector representation of the environment to an action for the vehicle dynamics with an attention-based neural network architecture. This model is optimized with Proximal Policy Optimization (PPO) \cite{schulman2017proximal}. Subsequently, we collect continuous driving data from 15 diverse virtual environments with randomly generated traffic conditions. Our collection includes a 100k dataset for pretraining, a 10k set for QA labeling and fine-tuning, and a 1k set dedicated to evaluation.

\subsection{Structured Language Generation for Pseudo Vector Captioning}
\label{vector_cap}
In our framework, we aim to convert vector representations into language using a structured language generator to facilitate the grounding the vector representation into LLMs. Since our object-level vectors contain semantically significant attributes, such as the number of cars and pedestrians, their respective locations, orientations, speeds, bounding boxes and other attributes, we employ a structured language generator (lanGen) function to craft pseudo-language labels derived from the vector space, as illustrated below:

\begin{align*}
&\text{lanGen}(v_{\text{car}},v_{\text{ped}},v_{\text{ego}},v_{\text{route}},[o_{rl}]) \to \\
&\left\{\begin{array}{p{0.8\columnwidth}}
\scalebox{2}{``}A moving car; Angle in degrees: 1.19; Distance: 9.98m; [My attention: 78\%]\\
A pedestrian; Angle in degrees: -41.90; Distance: 11.94m; [My attention: 22\%]\\
My current speed is 11.96 mph.\\
There is a traffic light and it is red. It is 12.63m ahead.\\
The next turn is 58 degrees right in 14.51m.\\[1ex]
[Here are my actions:]\\[0.1ex]
[  - Accelerator pedal 0\%]\\[0.1ex]
[  - Brake pedal 80\%]\\[0.1ex]
[  - Steering straight]\\[0.1ex]
\scalebox{2}{''}
\end{array}\right.
\end{align*}

In this function, variables $v_{\text{car}}$, $v_{\text{ped}}$, $v_{\text{ego}}$, and $v_{\text{route}}$ denote vector information corresponding to cars, pedestrians, ego vehicle, and route, respectively. The optional term $o_{\text{rl}}$ represents the output from the RL agent, consisting of additional attention and action labels for guiding the action reasoning process. Attention labels are collected from RL policy attention layers similar to \cite{renz2022plant}.

This lanGen enables the transformation of vector representations into human-readable language captions. It crafts a comprehensive narrative of the current driving scenario, which includes of the agent's observations, the agent's current state, and its planned actions. This comprehensive contextual foundation enables the LLMs to conduct reasoning and construct appropriate responses in a manner that humans can interpret and understand.

The inclusion of $o_{\text{rl}}$ is optional, and we generate two different versions of pseudo labels to cater to different requirements: 1) \textbf{Without Attention/Action:} Employed during the representation pre-training stage (see Subsection \ref{subsec:pretraining}), where the inference of attentions and actions is not required.
2) \textbf{With Attention/Action:} Utilized for VQA labeling with GPT during the fine-tuning stage (see Subsection \ref{subsec:training_smart_agent}). This equips GPT with the ability to ask specific questions about attentions and actions, thereby empowering the driving LLM agent with the ability to reason about attentions and actions.

\subsection{Driving QA Dataset Labeling}
\label{DQA}
Large amount of data is the key to enabling the question answering ability of the language models. This becomes particularly crucial when a new modality is introduced to the LLMs; in such circumstances, it is essential to have a high-quality question-answering dataset relevant to the modality input. Studies have demonstrated that data labelled using ChatGPT surpasses crowdsourced workers in terms of performance for text-annotation tasks \cite{gilardi2023chatgpt}. Inspired by Self-Instruct \cite{wang2022selfinstruct} and LLaVA \cite{liu2023visual}, we utilize GPT to generate a Driving QA dataset. This dataset is conditioned on the structured language input, serving as a representative for the vectors.

To be specific, we tasked GPT to emulate the role of a certified professional driving instructor to generate questions and answers for students. We supplied basic driving rules as pre-prompts, covering aspects like traffic signal transitions and the driving side as we are operating in a left-hand driving country. We tasked GPT with creating 16 sets of random questions and answers per scenario, taking into account the current driving scenario provided by language description generated by the lanGen. We also included some instructions to guide the generation of action-related questions, so that they can be easily parsed with regular expressions for evaluating action prediction tasks.

To reduce hallucinations, we established simple guardrail rules. We instructed GPT to include a driving-related question that could not be inferred from the observation input. If faced with such a question, the model's response should be akin to "I'm unable to answer this question based on the observations I have," preventing the model from imagining objects not present in the provided vector observation.

Additionally, we included a random question unrelated to driving, instructing the model to respond with "As an AI Driver, the question you asked is out of my scope, but I can attempt to answer it." The model would then proceed to answer the question normally, thereby preserving the language model's versatility as a general-purpose assistant.

Detailed instructions can be found in our open-sourced code\footref{code}. Below are some examples of the question-answer pairs generated by GPT:
\begin{itemize}
\item \textbf{Question:} "What objects are you observing?" \
\textbf{Answer:} "I'm observing a moving car, and two pedestrians. The car is coming from the opposite direction, and the pedestrians are crossing the street from both the right and left sides."
\item \textbf{Question:} "How are you going to drive in this situation and why?" \ \textbf{Answer:} "Here are my actions: - Accelerator pedal 0\% - Brake pedal 7\% - Steering straight. Reason: Because I'm maintaining a slow speed due to the pedestrians crossing, and I'm steering straight to keep our vehicle properly aligned with the road."
\item \textbf{Question:} "What are the best tourist spots in London?" \
\textbf{Answer:} "As an AI Driver, the question you asked is out of my scope, but I can try to answer it. Some famous tourist spots in London include the Tower of London, Buckingham Palace, The British Museum, The Shard, and the London Eye."
\end{itemize}

\subsection{Training the Driving LLM Agent}

\begin{figure*}[hbt]
    \centering
    \includegraphics[width=1.0\linewidth]{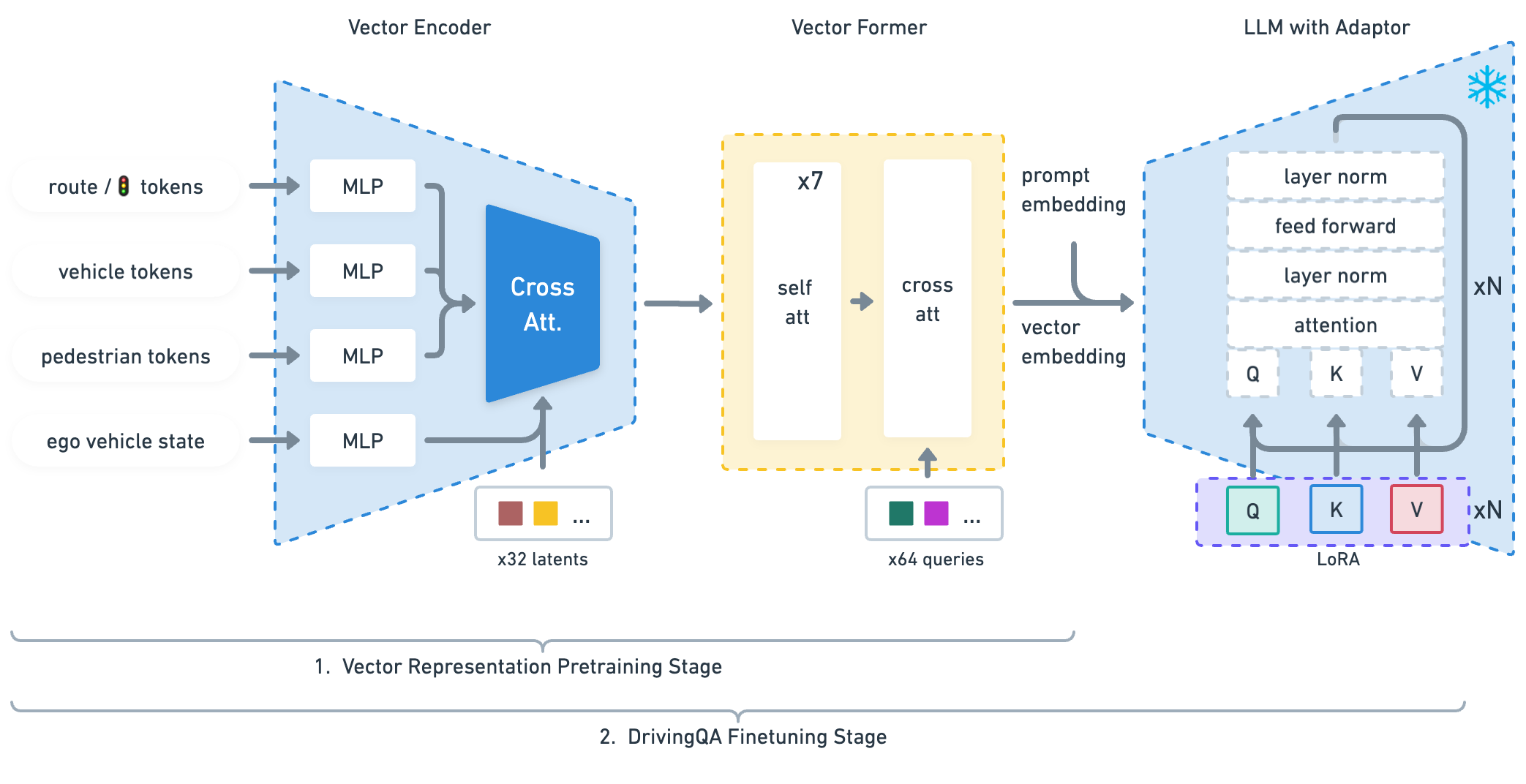}
    \caption{The architecture of the Driving LLM Agent} 
    \label{fig:arc} 
\end{figure*}

Training the LLM-Driver involves formulating it as a Driving Question Answering (DQA) problem within the context of a language model. The key to this formulation is the integration of an object-level vector modality into the pre-trained LLMs, creating a multi-modal system capable of interpreting and interacting with both language and vector inputs.

We use a two-stage process to train our model for effectively fusing object-level vector modality into the LLM-Driver. In the first stage, we ground the vector representation into an embedding that can be decoded by the LLMs. This is accomplished by freezing the language model and optimizing the weights of the vector encoders and the vector transformer. In the second stage, we finetune the model to the DQA task, training it to answer driving-related questions and take appropriate actions based on its current understanding of the environment.

As can be seen in the Figure \ref{fig:arc}, our model is built on three key components: the Vector Encoder, Vector Former, and a frozen LLM with a Low-Rank Adaptation (LoRA) \cite{hu2021lora} module.
\begin{itemize}
\item \textbf{Vector Encoder}: The four input vectors are passed through the Multilayer Perceptron (MLP) layers. They're then processed by a cross-attention layer to move them into a latent space. We add the ego feature to each learned input latent vector to emphasize the ego states.
\item \textbf{Vector Former}:  This part contains self-attention layers and a cross-attention layer that work with the latent space and question tokens. This transforms the latent vectors into an embedding that the LLM can decode.
\item \textbf{LLM with Adaptor}: Here, we inject trainable rank decomposition matrices (LoRA) into the linear layers of the pretrained LLMs for parameter-efficient finetuning. We utilize LLaMA-7b \cite{touvron2023llama} as the pretrained LLM for our experiments.
\end{itemize}

\subsubsection{\textbf{Vector Representation Pre-training}}
\label{subsec:pretraining}
Integrating a new modality into pre-trained Large Language Models (LLMs) poses significant challenges due to the need for extensive data and computational resources. In this study, we propose a novel approach that leverages structured language to bridge the gap between the vector space and language embeddings, particularly focusing on numerical tokens.

During the pretraining phase, we freeze the language model while training the entire framework end-to-end to optimize the weights of the vector encoders and the vector transformer (V-former). Such an optimization process enables effective grounding of the vector representation into an embedding that can be directly decoded by the LLMs. It is important to note that during this pretraining phase, we use only perception structured-language labels and avoid training on tasks that involve reasoning, such as action prediction (vehicle control commands) and agent attention prediction (where the expert focuses spatial attention). This is because our focus at this stage is solely on representation training, and we aim to avoid prematurely integrating any reasoning components into the V-former.

The pretraining process was conducted using 100k question-answer pairs, which were collected from a simulator. Additionally, in each epoch, we sampled 200k vectors with uniformly-distributed random values, contributed to robust representation learning by comprehensively exploring the vector space and its associated semantic meanings. We employed the lanGen method, as described in Sections \ref{vector_cap}, to automatically label the pseudo vector-captioning data. During the pretraining phase, we penalized errors in vector captioning results to optimize the vector encoder and V-former weights, thereby transforming the vector space into language embeddings understandable by LLM.

Through this approach, we are able to effectively incorporate object-level vector modality into pre-trained LLMs, which serves as a good starting point for the finetuning stage.

\subsubsection{\textbf{Driving QA Finetuning}}
\label{subsec:training_smart_agent}
After the pre-training stage, we integrate the trainable LoRA module into the LLM, and optimize the weights of the Vector Encoder, Vector Former and LoRA module in an end-to-end fashion on the Driving QA data that we collected in Section \ref{DQA}.

In order to train the LLM-Driver to output accurate actions for driving, we added certain action-triggering questions to the VQA dataset. These are questions that, when asked, require the agent to generate actions in specific format. For example, a question like "How are you going to drive in this case and why?" would require the agent to infer the action based on the vector input. We then use a simple language-action grounding strategy with regular expression matching to extract the action required to control the car in our simulator. To ensure the model pays sufficient attention to these important questions, we upsample the action-triggering questions with different expressions such as:

\begin{itemize}
\item "How are you going to drive in this situation?"
\item "What actions are you taking?"
\item "How are you driving in this situation?"
\item "What are your planned actions for this situation?"
\end{itemize}

We refine the pre-existing model on 10,000 scenarios, each incorporating 16 driving QA entities, along with an additional vector for captioning data and an action prediction QA pair. The outcome is a driving LLM agent that is capable of reasoning about its actions and responding to any driving-related questions given the object-level vectors of the current driving scenario.

\section{Evaluation and Results}
This section aims to thoroughly assess our model's performance using a separate evaluation set comprising 1,000 distinct driving scenarios. The vector captioning and vector QA data for these scenarios were labeled following the same methodologies discussed in Sections \ref{vector_cap} and \ref{DQA}.

\subsection{Evaluation of Perception and Action Prediction}

\begin{figure*}[hbt]
    \centering
    \includegraphics[width=0.9\linewidth]{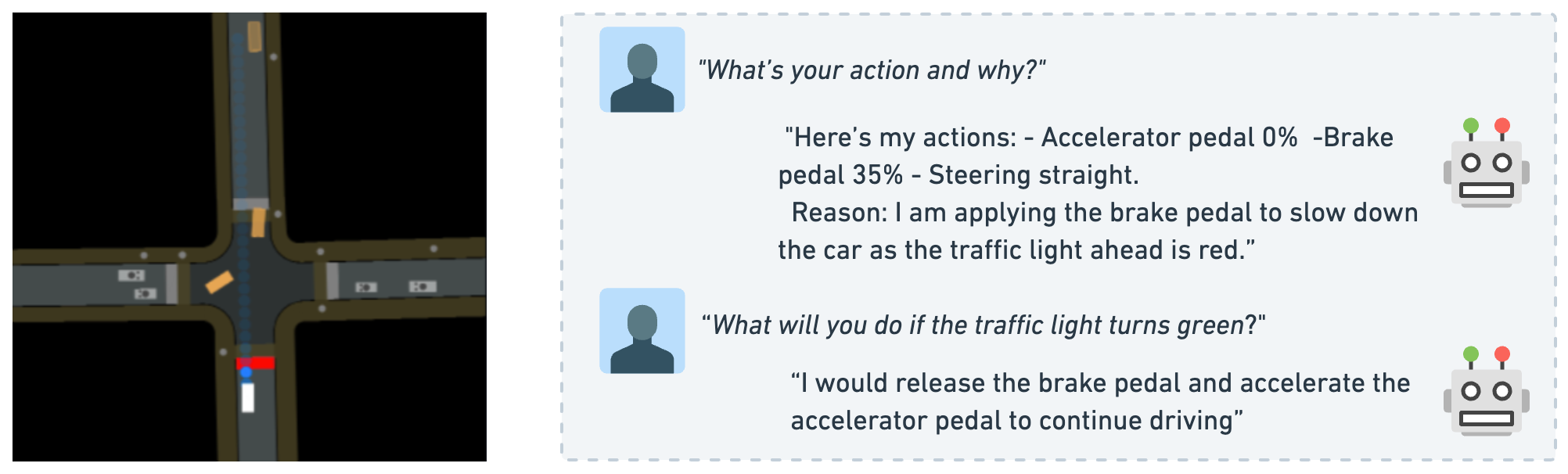}
    \caption{Demonstration of the interaction with the pretrained LLM-Driver. Notably, the driving LLM agent has the ability to anticipate future actions when prompted. Please refer to Appendix B for additional results, including some failure cases.} 
    \label{fig:tl} 
\end{figure*}

To obtain the model's perception and action predictions, we prompt the model with the same vector captioning and add the action-predicting triggering question from the driving QA data. We then parse the language output using regular expressions to extract the numerical predictions.

We reported the results on the model trained using a two stage approach (with pretraining), and the model training only on driving QA dataset (without pretraining). We also include a baseline model, Perceiver-BC, which replaces the LLM with a non-pretrained Perceiver model for comparison purposes. This aims to evaluate how effectively pretrained LLMs can be utilized for reasoning in driving-related tasks, such as action prediction. The Perceiver-BC model includes the identical Vector Encoder and V-former modules as those in the LLM agent model. However, it differs in the fact that it employs a transformer-based policy module in place of the LLM with adapters (please refer to Appendix A for more details). The Perceiver-BC model also outputs actions with perception auxiliary tasks of agent and traffic light detection. To maintain an equitable comparison, we've calibrated the BC model to have a similar number of trainable parameters as in the LLM agent, totaling to approximately 25 million trainable parameters. The Perceiver-BC model was trained on perception and action prediction tasks using the same 10k dataset that we used for the LLM Agent, but without VQA data. 

For the reported metrics, we calculate the Mean Absolute Error (MAE) for the predictions of the number of cars and pedestrians, denoted as $E_{car}$ and $E_{ped}$ respectively. Additionally, we measure the accuracy of traffic light detection as well as the mean absolute distance error in meters for traffic light distance prediction, represented as $Acc_{TL}$ and $D_{TL}$. Furthermore, we compute the MAE for normalized acceleration and brake pressure denoted as $E_{lon.}$, and normalized steering wheel angle denoted as $E_{lat.}$. Lastly, we report the weighted cross-entropy loss for the token prediction on the evaluation set, indicated as $L_{token}$.

\renewcommand{\arraystretch}{1.2} 

\begin{table*}[htb]
\centering
\caption{The evaluation result of perception and action prediction}
\begin{threeparttable}
\begin{tabular}{lcccccccccc}
\toprule
 & \multicolumn{2}{c}{\textit{\small agents count}} & \multicolumn{2}{c}{\textit{\small traffic light}} & \multicolumn{2}{c}{\textit{\small action}}  & {\textit{\small loss}} \\ 
\cmidrule(lr){2-3} \cmidrule(lr){4-5} \cmidrule(lr){6-7} \cmidrule(lr){8-8} 
& $\small E_{car}\downarrow$ & $\small E_{ped}\downarrow$ & $\small Acc_{TL}\uparrow$ & $\small D_{TL}\downarrow$ & $\small E_{lon.}\downarrow$ &$\small E_{lat.}\downarrow$ & $\small L_{token}\downarrow$ \\
\midrule
Perceiver-BC\cite{jaegle2021perceiver} &0.869& 0.684& \textbf{0.900}& \textbf{0.410}& 0.180& 0.111 & n/a  \\
\midrule
LLM-Driver$_{\text{w/o pretrain}}$ &0.101& 1.668& 0.758& 7.475& 0.094& 0.014\tnote{a} & 0.644  \\
LLM-Driver$_{\text{w/ pretrain}}$ &\textbf{0.066}& \textbf{0.313}& 0.718& 6.624& \textbf{0.066}& \textbf{0.014}\tnote{b}  &\textbf{0.502}\\
\bottomrule
\end{tabular}
\begin{tablenotes}
    \item[a] Exact value: 0.01441
    \item[b] Exact value: 0.01437
\end{tablenotes}
\end{threeparttable}
\label{result_table}
\end{table*}

As can be seen in Table \ref{result_table}, the results clearly demonstrate that the pretraining stage significantly enhances both the model's perception and action prediction capabilities. This suggests that the pretrained model exhibits a higher level of accuracy in perceiving and quantifying the number of cars and pedestrians in its environment. The pretrained model also shows a lower loss value, $L_{token}$, which indicates an improvement in the overall effectiveness of the model's token predictions.

Note that we filter out agents that fall outside the 30m range from the ego vehicle, which needs to be calculated using the x, y, z vector. This setting makes the "direct decoding" of agent detection from the vector much more difficult. For simpler regression tasks (e.g., traffic light distance), Perceiver-BC performs much better than LLMs.

For the action prediction task requiring in-depth reasoning, we found that LLM-based policies outperform the Perceiver-BC approach when given the same amount of training data and trainable parameters. This indicates that LLMs serve as effective action predictors, harnessing knowledge acquired during the general pretraining phase**—such as stopping at a red light or decelerating when vehicles or pedestrians are ahead—**to inform decisions based on their grounded observations. 

However, it's important to note the distinction in training methodologies: Perceiver-BC is trained mostly using regression on perception and control outputs, while LLMs are trained via cross-entropy token loss and benefit from an extra 16x pairs of driving questions and answers, which will reinforces the learning of perception and action prediction. Thus, the comparison might not be entirely equitable and should be taken as merely a point of reference.

\subsection{Evaluation of Driving QA}
To assess the quality of answers to open-ended questions about the driving environment, we use GPT-3.5 to grade our model's responses. This is a recently emerged technique for grading natural language answers \cite{fu2023gptscore} \cite{wang2023chatgpt} \cite{liu2023geval}. This approach allows us to quickly and consistently evaluate our model's capabilities for questions that don't have fixed answers.

For evaluation, we prompt GPT-3.5 with the language-based observation description used during dataset generation (section \ref{DQA}), the question from the test set, and the model's answer. GPT's task is to write a one-line assessment, followed by a score ranging from 0 to 10 (where 0 is worst and 10 is best) for each response, based on the given observation details. The final score of the model is the average of all question scores. We note that GPT evaluations can sometimes be overly lenient with answers that are incorrect but semantically close, as well as other issues with GPT-based grading as reported by \cite{liu2023geval}. To validate these findings, we hand-scored 230 randomly sampled QA pairs and obtained comparable results.

\begin{table}[htb]
\caption{\label{grading-results} Grading of the Driving QA outputs}
\begin{center}
\begin{tabular}{@{}lcc@{}}
\hline
& GPT Grading $\uparrow$& Human Grading $\uparrow$ \\
\midrule
Constant answer "I don't know"    & 2.92& 0.0\\
Randomly shuffled answers    & 3.88& 0.26\\
\midrule
$\text{LLM-Driver}_{\text{w/o pretrain}}$   & 7.48& 6.63\\
$\text{LLM-Driver}_{\text{w pretrain}}$ & \textbf{8.39}& \textbf{7.71}\\
\midrule
Answers generated by GPT       & 9.47& 9.24\\
\bottomrule
\end{tabular}

\end{center}
\label{grading}
\end{table}

Our results in Table \ref{grading} demonstrate that pre-training improves the grading score of the model by 9.1\% and 10.8\% over those models without pre-training, according to both GPT grading and human grading. For reference, we also provide the scores obtained when running our evaluation procedure with artificially incorrect answers (by either constantly answering "I don't know" or by randomly shuffling all answers), as well as the "ground-truth" answer labels provided by GPT. From these results, we can see that our model, which utilizes only vectorized representations as input, can achieve a much higher score than either a constant or random answer approach.

\section{Conclusion and Limitations}
While our approach exhibits considerable novelty and potential, the preliminary results indicate that there is still progress to be made for the LLM to fully navigate in simulation. We are aware of possible discrepancies in open-loop vs closed-loop results \cite{codevilla2018offline}, and our future work will focus on addressing the challenge of efficiently evaluating the model in a closed-loop system. This includes addressing the lengthy inference time of LLMs and the extensive steps needed to thoroughly test the system. Moreover, we anticipate the need to improve the precision of the direct driving commands produced by our baseline in order to operate effectively in closed-loop settings. Factors contributing to this include the intricacy of the task, potential enhancements to the model architecture, and the need for improving the scale and quality of our training dataset. We have observed that numeric inaccuracies and the early developmental stage of our system can lead to discrepancies in explanations and reasoning, preventing our ideas from being fully implemented in closed-loop settings. These observations will guide our future research as we continue to refine our approach.

Nevertheless, our work establishes a foundation for future research in this direction. We believe that our proposed architecture, combined with the novel way for grounding a new modality into LLMs, a data auto-labelling pipeline, and an LLM-based evaluation pipeline, can serve as a starting point for researchers interested in exploring the integration of numeric vector modality with LLMs in the context of autonomous driving.

In terms of future research, improving the architecture to better handle nuances in the numeric vector modality could offer a promising direction. Our approach holds potential for application beyond simulated environments. Given sufficient real-world perception labels for pretraining and fine-tuning a VLM, our methodology could also be adapted to real-world driving scenarios.

Overall, while our results are preliminary we believe our work is a significant step forward in the integration of vector modality with LLMs in the context of autonomous driving. 

\section{Ethical Implications and Broader Impact}

By introducing LLMs into autonomous driving we are inheriting their ethical implications \cite{zhuo2023exploring}. We believe that these problems will be addressed by further research from the LLM community. As for autonomous driving, it's crucial to ensure that the system can handle all possible driving scenarios safely. By improving the interpretability of autonomous driving systems, we can help build trust in this technology, thereby accelerating its adoption and ultimately leading to safer and more efficient transportation systems.

\bibliographystyle{IEEEtran}
\bibliography{IEEEabrv,references.bib}

\clearpage
\begin{appendices}
\section{Perceiver-BC model}
\begin{figure}[hbt]
    \centering
    \includegraphics[width=\linewidth]{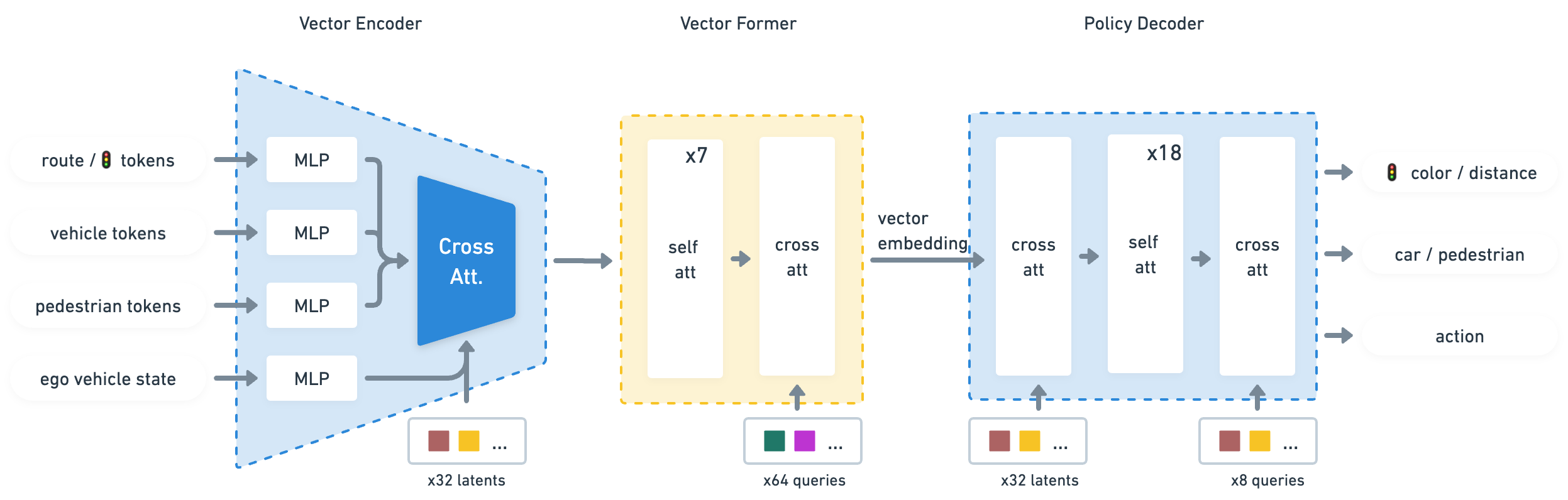}
    \caption{The architecture of the behaviour cloning baseline} 
    \label{fig:bc} 
\end{figure}

In order to assess the proficiency of leveraging the reasoning ability of pretrained Large Language Models (LLMs) in driving tasks, such as action prediction, we decided to conduct additional experiments using a simple Perceiver-BC model for comparison purposes. We've detailed the implementation process in Figure \ref{fig:bc}.

The Perceiver-BC model includes the identical Vector Encoder and Vector Former modules as those in the LLM agent model. However, it differs in that it employs a perceiver-based policy module instead of the LLM with adapters. To maintain an equitable comparison, we've calibrated the BC model to possess a similar number of trainable parameters to the LLM agent, totaling approximately 25 million trainable parameters.

The Perceiver-BC model was trained using the same 10k dataset that we used for the LLM Agent. This dataset included only captioning and action data. We trained the Perceiver-BC model for 5 epochs, just like the LLM Agent. We have reported the results of these evaluations in Table \ref{result_table}.

 \section{Additional Results}
\begin{figure}[hbt]
    \centering
    \includegraphics[width=\linewidth]{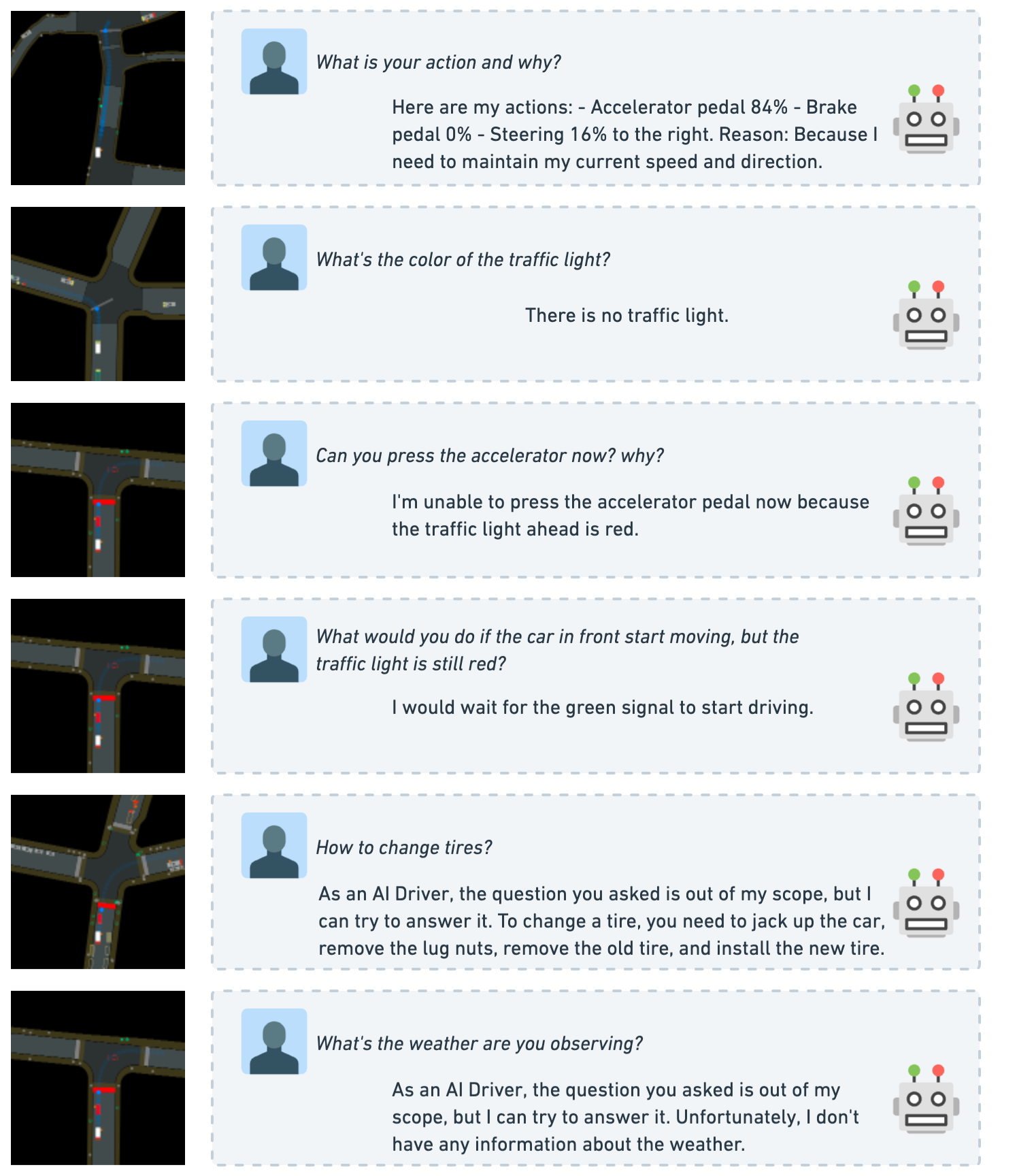}
    \caption{More drivingQA results} 
    \label{fig:results} 
\end{figure}

\begin{figure}[hbt]
    \centering
    \includegraphics[width=\linewidth]{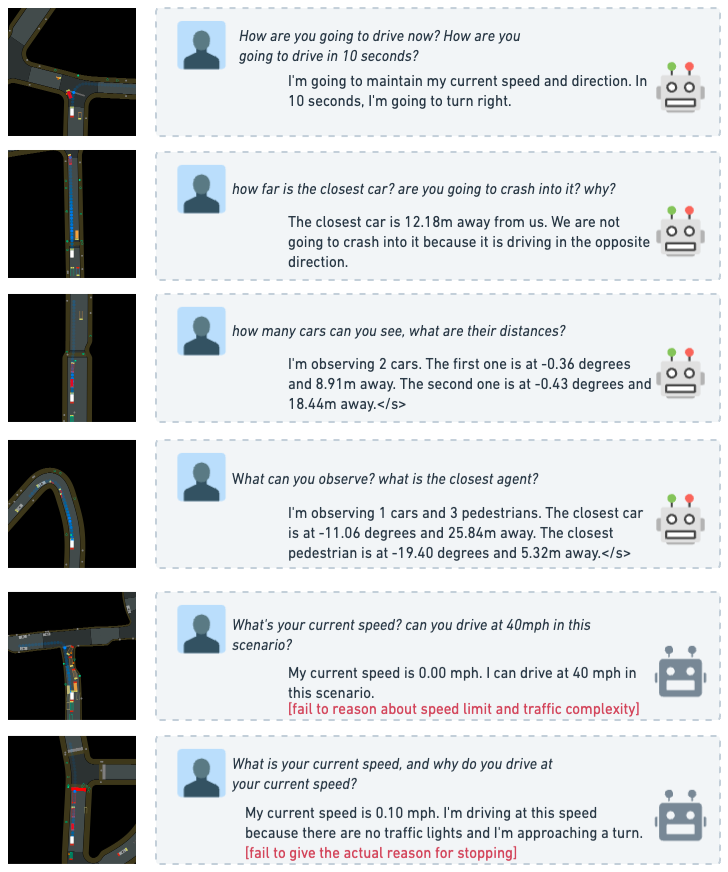}
    \caption{More drivingQA results and failure cases (last two)} 
    \label{fig:results1} 
\end{figure}

We include here more results DrivingQA results in Figure \ref{fig:results} and Figure \ref{fig:results1}, including some failure cases.

\clearpage
\subsection{Design of the Object-level Vector Representation} \label{sec:object-level-repr}

In order to effectively translate the intricacies of autonomous driving into a format that a Large Language Model (LLM) can understand and work with, we have designed an object-level vector representation. This allows us to represent a variety of crucial aspects of driving conditions, such as the intended route, nearby vehicles, pedestrian presence, and the state of the ego vehicle, using a set of compact object-level vectors.

Our object-level vectors contain semantically significant attributes. This allows us to further create structured language labels from the vector representation. This process turns complex driving situations into a language format that the Large Language Model (LLM) can understand. This enables the LLM to make sense of driving conditions, make decisions, and even generate outputs that can be explained, based on its own internal reasoning.

The schema below provides more details about each aspect represented in our vector input system:

\begin{minted}{python}
    # A 2d array per ego vehicle describing the route to follow.
    # It finds route points for each vehicle. Each point goes into a new row, 
    # then each point have the following attributes:
    # - (x, y, z) of a point
    # - its direction
    # - pitch
    # - speed limit
    # - is junction?
    # - width of the road
    # - is traffic light and its state
    # - is give way?
    # - is a roundabout?
    route_descriptors: torch.FloatTensor

    # A 2d array per ego vehicle describing nearby vehicles.
    # First, if finds nearby vehicle in the neighbourhood of the car.
    # Then allocates an array of zeros a fixed max size (about 30).
    # There is a logic that tries to fit dynamic and static vehicles into
    # rows of that array.
    # For every vehicle:
    # - "1" for marking that a vehicle is found in the row 
    # (empty rows will have this still "0")
    # - Is it dynamic or static (parked) vehicle
    # - its speed
    # - its relative position in the ego coordinates
    # - its relative orientation
    # - its pitch
    # - its size
    # - vehicle class
    # - positions of its 4 corners
    vehicle_descriptors: torch.FloatTensor

    # A 2d array per ego vehicle describing pedestrians.
    # First, if finds nearby pedestrians in the neighbourhood of the car.
    # Then allocates an array of zeros a fixed max size (about 20).
    # Then every found pedestrian is described in a row of that array:
    # - "1" for marking that a pedestrian is found in the row
    # (empty rows will have this still "0")
    # - ped. speed
    # - its relative position in the ego coordinates (x, y, z)
    # - its relative orientation
    # - pedestrian type
    # - intent of crossing the road
    pedestrian_descriptors: torch.FloatTensor

    # A 1D array per ego vehicle describing its state. Specifically,
    # - VehicleDynamicsState (acc, steering, pitch ..)
    # - Vehicle size
    # - Vehicle class
    # - Vehicle dynamics type
    # - Previous action
    # - 2 lidar distance arrays, placed on the front corner of the vehicle
    ego_vehicle_descriptor: torch.FloatTensor
\end{minted}

\subsection{RL expert}
In this section we describe how we trained the reinforcement learning agent used to collect data in a diverse set of driving scenarios. 

The agent maps a set of object-level vectors, described in section \ref{sec:object-level-repr}, to actions executable by the simulated vehicle dynamics. The actions of the RL agent are modelled by two independent beta distributions, corresponding to acceleration and steering wheel angle. We partition the acceleration value $x \in [0; 1]$, sampled from the beta distribution, into accelerator pedal and brake pressure:

\begin{align*}
    \text{Accel} = s_1 \max\{0; 2x - 1\}\\
    \text{BrakePressure} = s_2 \max\{0; 1 - 2x\}
\end{align*}

where $s_1$ and $s_2$ scale the values to the respective output ranges. The three values accelerator pedal, brake pressure, and steering wheel angle are then fed into our vehicle dynamics simulation that closely matches our real world vehicles.

To learn the action distributions, we chose Proximal Policy Optimization \cite{schulman2017proximal} as a reinforcement learning algorithm due to its simplicity, widespread usage, and robustness. In each epoch of training we run the simulation with the agent in closed-loop and obtain a replay buffer. Note that we control multiple vehicles simultaneously, which allows us to generate a replay buffer with 100 to 200 thousand samples very efficiently. We then retrieve batches from that replay buffer and optimize the neural network using the PPO objective. After that, the replay buffer is discarded, and we begin the next epoch. We run this procedure for 10 thousand steps. 

As the neural architecture, we use Perceiver IO \cite{jaegle2021perceiver}. The advantage of this architecture lies in its flexibility to deal with different input and output types. Input vectors from the simulator are preprocessed into tokens using multi-layer perceptrons with ReLU non-linearity, which are then fed into the cross-attention mechanism. Two output queries of the Perceiver are used to predict acceleration and steering wheel distributions, respectively. An additional output query predicts a value estimate.

The reward during training is given by the following formula, which is based on the rewards described in \cite{toromanoff2020end}.

\begin{align}
    R = \underbrace{\beta_1 \left(1 - \frac{|v - v_{\mathrm{desired}}|}{v_s}\right)}_{\text{Desired speed}} - \underbrace{\beta_2 d_{\mathrm{lateral}}}_{\text{Lane distance}} - \underbrace{\beta_3 \alpha_{\mathrm{angle}}}_{\text{Lane angle}} - \underbrace{\beta_4\frac{\left(\phi_t - \phi_{t-1}\right)^2}{\phi_s}}_{\text{Erratic steering penalty}} + \underbrace{R_{\text{failed to stop}}}_{\text{Stopping penalty}}
\end{align}

where $v$ is the vehicle speed, $v_{\mathrm{desired}}$ is a desired speed heuristic (described below), $d_{\mathrm{lateral}}$ is the distance from vehicle centre to closest lane centre, $\alpha_{\mathrm{angle}}$ is the angle between vehicle and closest lane tangent, $\phi_t$ and $\phi_{t-1}$ are steering wheel angle actions from the current and previous time step, and $R_{\text{failed to stop}}$ is another heuristic for yielding the right-of-way. $\beta_1$, $\beta_2$, $\beta_3$, and $\beta_4$ are hyper-parameters and $v_s$ and $\phi_s$ are velocity and steering wheel scales.

The desired speed $v_{\mathrm{desired}}$ serves as a heuristic for determining the optimal speed in various situations. We address cases where the agent is in a junction, near another vehicle, at a traffic light, close to a pedestrian, or trailing a lead vehicle. For each of these situations, we establish a heuristic, and the final desired speed is selected as the minimum among them.

The failed to stop heuristic ensures the agent stops appropriately. It is given by

\begin{align}
R_{\text{failed to stop}} = \begin{cases}
    -v & \text{if the agent is close to a give way sign and $v > 2m/s$}\\
    0 & \text{otherwise}
\end{cases}
\end{align}

\end{appendices}

\end{document}